\documentclass[conference]{IEEEtran}
\usepackage{blindtext, graphicx, amsmath, caption, subfigure}

\begin{document}
%
\title{Exploring Geometric Property Thresholds For Filtering Non-Text Regions In A Connected Component Based Text Detection Application}

\author{\IEEEauthorblockN{Teresa Nicole Brooks}
\IEEEauthorblockA{Seidenberg School of CSIS \\Pace University
New York, NY\\
tb93141n@pace.edu}}

\maketitle

\IEEEpeerreviewmaketitle

\begin{abstract}
Automated text detection is a difficult computer vision task. In order to accurately detect and identity text in an image or video, two major problems must be addressed. The primary problem is implementing a robust and reliable method for distinguishing text vs non-text regions in images and videos. Part of the difficulty stems from the almost unlimited combinations of fonts, lighting conditions, distortions, and other variations that can be found in images and videos. This paper explores key properties of two popular and proven methods for implementing text detection; maximum stable external regions (MSER) and stroke width variation.
\end{abstract}

\section{Introduction}
Automated text detection is employed in modern software systems to perform tasks such as detecting landmarks in images and video, surveillance, visual sign translation mobile apps and in more recent years self-driving cars. Many of these applications require a robust automated text detection system that can accurately detect text on signs in natural and unnatural images. 

One popular and proven approach to text detection is using connected components (CC) \cite{Neumann2015a} \cite{Chen2011a} to detect text. In connected component based systems the goal is to separate non-text and text region candidates. This is done by using learned or manually established features (thresholds) for discriminating geometric properties that are used to filter out non-text regions. One of the most popular feature detectors used in conjunction in connected component based systems is Maximally Stable Extermal Regions (MSER) \cite{Bergasa2012a} \cite{Chen2011a}. 

The geometric property thresholds selected for filtering non-text regions can vary from image to image and are often affected by characteristics such as font style, image distortions (blur, skew etc) and textures to name a few. The objectives of this research is to explore and learn the effects of different font styles in both natural scene and unnatural scene images such as product boxes (i.e. cereal boxes) have on selecting these geometric property thresholds and their implications for the robustness of text detection in images. This research will also explore which geometric properties work best across both natural and unnatural scene images. 

\section{Background}
This section discusses the techniques and algorithms that are used in the exploratory text detection application implemented for this research project.

\subsection{Maximally Stable Extermal Regions (MSER)}
The exploratory text detection application implemented for this project uses Maximally Stable Extermal Regions (MSER) to perform the initial feature (region) detection in our test images. MSER is a popular feature detection algorithm that detects stable, connected component regions of gray level images called MSERs. This algorithm was first introduced by Matas et al. This paper introduced the use of MSER detection to solve the ``wide-baseline stereo problem \cite{Extremal2002}.'' This problem seeks to find the correspondence between elements in two images that were taken at different view points.  One major contribution of this paper is the introduction of ``extermal regions'' \cite{Extremal2002}. Extermal regions are regions that are maximally stable, where stability is defined by lack of or very low variation in intensity values with respect to some established thresholds.  Though MSER is sensitive to noise and blur \cite{Republic}
it offers important advantages over other region based detection algorithms as it offers affine invariance, stability, and highly repeatable results \cite{Mikolajczyk2005}.

\subsubsection{Region Geometric Properties}
Once connected components regions (MSERs) are detected, we can use certain discriminating, geometric properties to filter out non-text region candidates. The regions are filtered out via the use of thresholds. The property values are features whereby thresholds can be manually established, calculated or learned using machine learning techniques.  These features are a fast and easy way to discern non-text regions from  text features in images \cite{Bergasa2012a}. Below is a brief discussion of the discriminating, geometric properties used in the system implemented for this project.

\begin{itemize}
  \item \textbf{Aspect ratio:} The ratio of the width to the height of bounding boxes.
  \item \textbf{Eccentricity:} Is used to measure the circular nature of a given region. It is defined as the distance between the foci and/or the ellipse and its major axis.
  \item \textbf{Solidity:} Is ratio of the pixels in the convex hull area that are also in a given region. It is calculated by \cite{Bergasa2012a} \[\frac{\text{area}}{\text{convex area}}\] 
  \item \textbf{Extent:} The size and location of the rectangle that encloses text.
  \item \textbf{Euler Number:} Is a feature of a binary image. It is the number of connected components minus the number of holes (in those components) \cite{Republic2012a}. 
\end{itemize}


\subsection{Stroke Width Variation}
Stroke width is another property that is used to distinguish (and filter) non-text regions from text-regions. The approach of using stroke width to aid in detecting text regions in images was introduced by Epshtein et al \cite{Epshtein2010}. The main idea behind using stroke width to filter non-text region is based around on the fact that usually text vs other elements in an image have constant stroke width. This means that binary images can be transformed into stroke width images (skeleton images) and these skeletal images are used to calculate the stroke width variation. This ratio can be used with a max variance threshold to filter non-text regions. Because stroke widths are calculated on a per pixel basis text detection systems that use this technique can detect text in a way that is insensitive to font, color, size, language and orientation \cite{Li2012a}. Below is a brief discussion of the stroke width related values that are calculated in the text detection system implemented for this project.

\begin{itemize}
  \item Stroke Width Variation Metric: \[\frac{\text{Standard Deviation of Stroke Width Values}}{\text{Mean of Stroke Width Values}}\]  \ 
  \item Stroke Width Max Threshold
\end{itemize}

\section{Related Work}
The literature reflects the impact and popularity of the use of Maximally Stable Extermal Regions (MSER) and Stroke Width Variation for detecting text in images. This section briefly discusses some key related works that influenced the approach taken in the application implemented for this paper.

\subsection{Text Location in Complex Images (2012)}
Gonzalez et al, presented a three stage text detection system. This system combined ``fast-to-compute'' \cite{Bergasa2012a} connected component features that were used to filter non-text regions, segmentation to detect text character candidates, and a support vector machine based text line classifier. Their segmentation approach combines the use of MSER and "locally adaptive threshold" \cite{Bergasa2012a} methods and their "fast-to-computer" features includes discriminating geometric properties such as aspect ration of bounding boxes, stroke width ration (metric) and solidity amount others. This approach proved to be especially effective in detected text in images with complex backgrounds. Their system it outperformed various systems when tested against the Robust Reading competitions datasets for IDCAR 2003, 2011 as well as the CoverDB test dataset. 

\subsection{Robust Text Detection In Natural Images With Edge-enhanced Maximally Stable Extremal Regions (2011)}
Chen et al, proposed a robust connected component based text detection system that implements an ``edge-enhanced'' \cite{Chen2011a} version of Maximally Stable Extermal Regions to find letter candidates. The enhanced version of MSER leveraged Canny edges, a robust edge detector operator in order to support detecting small text in low resolution images. This approach also enabled their system to mitigate MSER\'s known issues with detecting features in blurred images. This system also employed stroke width information to filter out non-text regions. They evaluated their system against the ICDAR competition dataset, and it performed well as or better than state of the art systems at the time the paper was published.

\section{Algorithm}
The text detection applications created for this research project implements the following workflow. This workflow performs pre-processing, text detection in a given image and optical character recognition when possible.

\begin{enumerate}
  \item Pre-process images. This includes, converting images to high contrast images, converting color images to gray scale images and rotating images (or of boudning boxes themselves) as needed.
  \item Process each image with MSER feature detector algorithm.
  \item Perform first pass of non-text region filtering on the features detected by MSER algorithm using discriminating geometric properties.
  \item Perform second pass of non-text region filtering on the remaining features using stroke width variation filtering.
  \item On the remaining regions (text regions), calculate the overlapping bounding boxes of these regions and combine them to create one bounded region of text.
  \item Detect the text in the image and pass the detected text bounding boxes to an ORC algorithm to determine the text that was found in the image. 
\end{enumerate}

\section{Experiments}
The primary goal of all experiments was to compare and contrast the discriminating geometric properties and their thresholds that worked well for detecting text in both natural and non-natural images. The secondary goal of these experiments was to understand the challenges if any of detecting text when an image contains highly stylized fonts.

The dataset consists of both natural images and non-natural images. The natural scene images in Fig 1 are from 2014 training dataset for the COCO Captioning Challenge competition. COCO is dataset for image recognition, segmentation and captioning \cite{coco2014}. The non-natural images in Fig 2 were taken for the purpose of this project.

\begin{figure}[!htb]
    \centering
    \captionsetup{justification=centering}
    \begin{subfigure}[Rotated Stop Sign]{\label{fig:a}\includegraphics[height=1.2in]{stop_on_main_small.jpg}}
    \end{subfigure}
    \begin{subfigure}[Simple Stop Sign]{\label{fig:b}\includegraphics[height=1.2in]{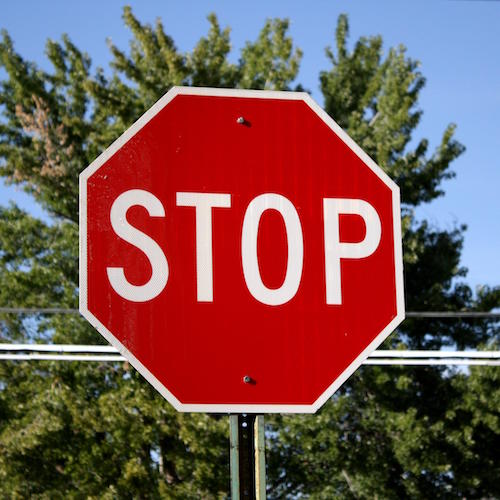}}
    \end{subfigure}
    \caption{Natural Images: Images from COCO Captioning Challenge 2014 training set.}
\end{figure}

\begin{figure}[!htb]
    \centering
    \captionsetup{justification=centering}
    \begin{subfigure}[Box Image 1]{\label{fig:a}\includegraphics[height=1.2in]{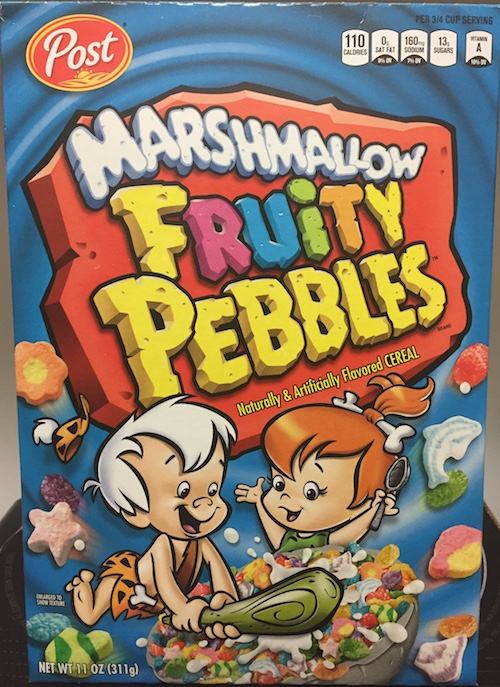}}
    \end{subfigure}
     \begin{subfigure}[Box Image 2]{\label{fig:b}\includegraphics[height=1.2in]{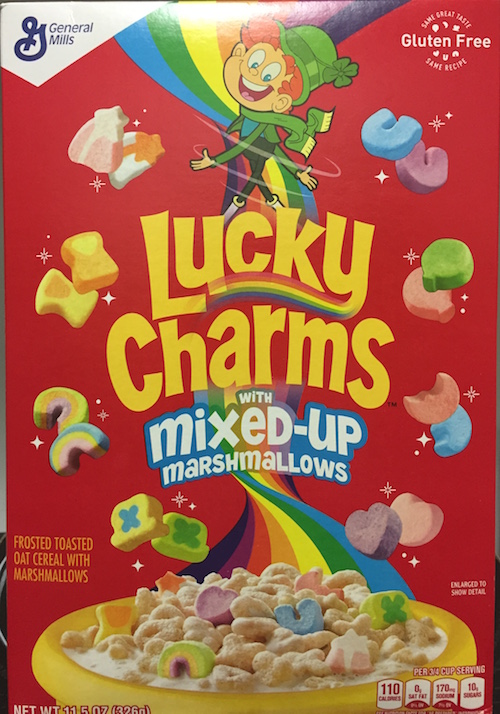}}
    \end{subfigure}
    \caption{Non-Natural Images: Cereal Boxes}
\end{figure}

The main objective of the trials conducted was to find the primary or dominate text regions in a given image. Primary or dominate text regions defined as the largest text region in the image. Each dataset image was processed individually by being run through the workflow presented in Section IV. During each iteration the geometric property thresholds, stroke width variation max threshold as well as other properties were manually tuned until the primary or dominate text region were detected. Note, all threshold and tuneable values started as the defaults defined by libraries used to implement the system.

\subsection{Object Character Recognition (OCR)}
The primary objective of all experiments was to visually verify text detection by plotting individual character and line or word bounding boxes on processed images. An additional verification step was employed using Matlab's OCR library. Matlab's OCR library is implemented by a popular, robust open source OCR engine called Tesseract \cite{tessORC}. Without any additional training the OCR library's default English language model was able to recognize the text from the natural image texts. One limitation encounter was that in order to recognize custom, highly stylized fonts one must training customer OCR models, because this was beyond the scope of the research project this task will be deferred for future work.

\section{Results}
The text detection application used in these experiments is not overly robust, but shows the power of the approach employed for text detection. Even with a modest implementation the application was able detect primary and in some cases secondary text in a given image with a few iterations of manually tuning thresholds, as Fig 3 and Fig 4 show.

\begin{figure}[!htb]
    \centering
    \captionsetup{justification=centering}
     \begin{subfigure}[]{\includegraphics[height=2in]{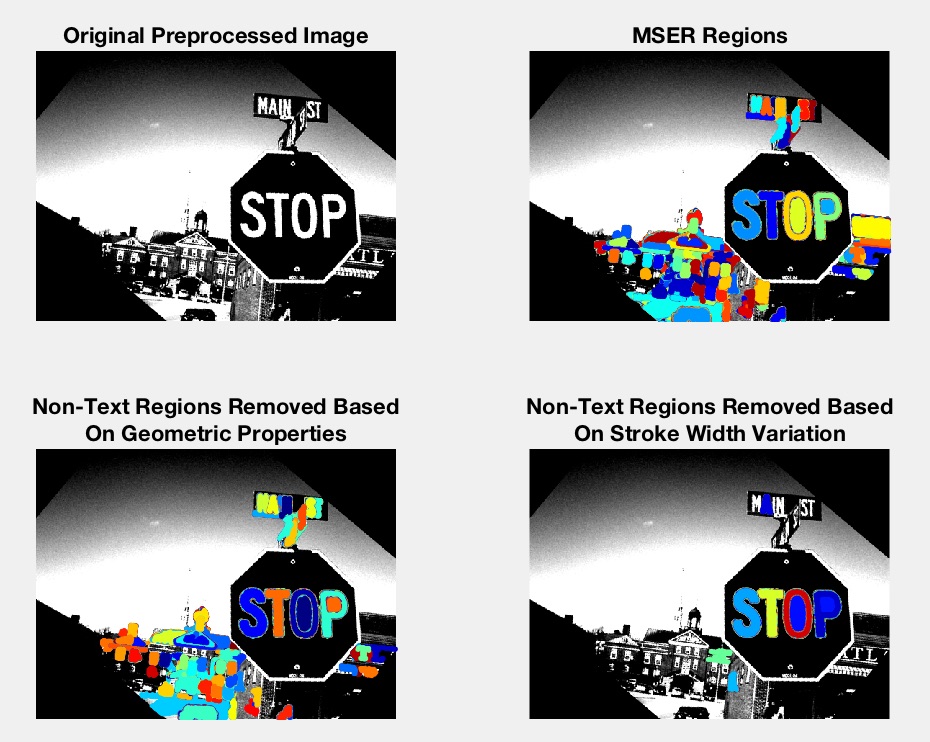}}
    \end{subfigure}
    \begin{subfigure}[]{\includegraphics[height=1.5in]{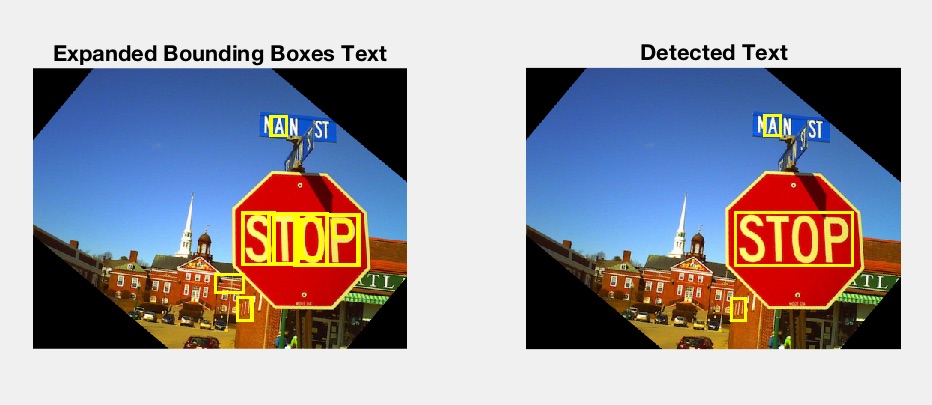}}
    \end{subfigure}
    \caption{Detected Features \& Text In Natural Image}
\end{figure}

\begin{figure}[!htb]
    \centering
    \captionsetup{justification=centering}
     \begin{subfigure}[]{\includegraphics[height=2in]{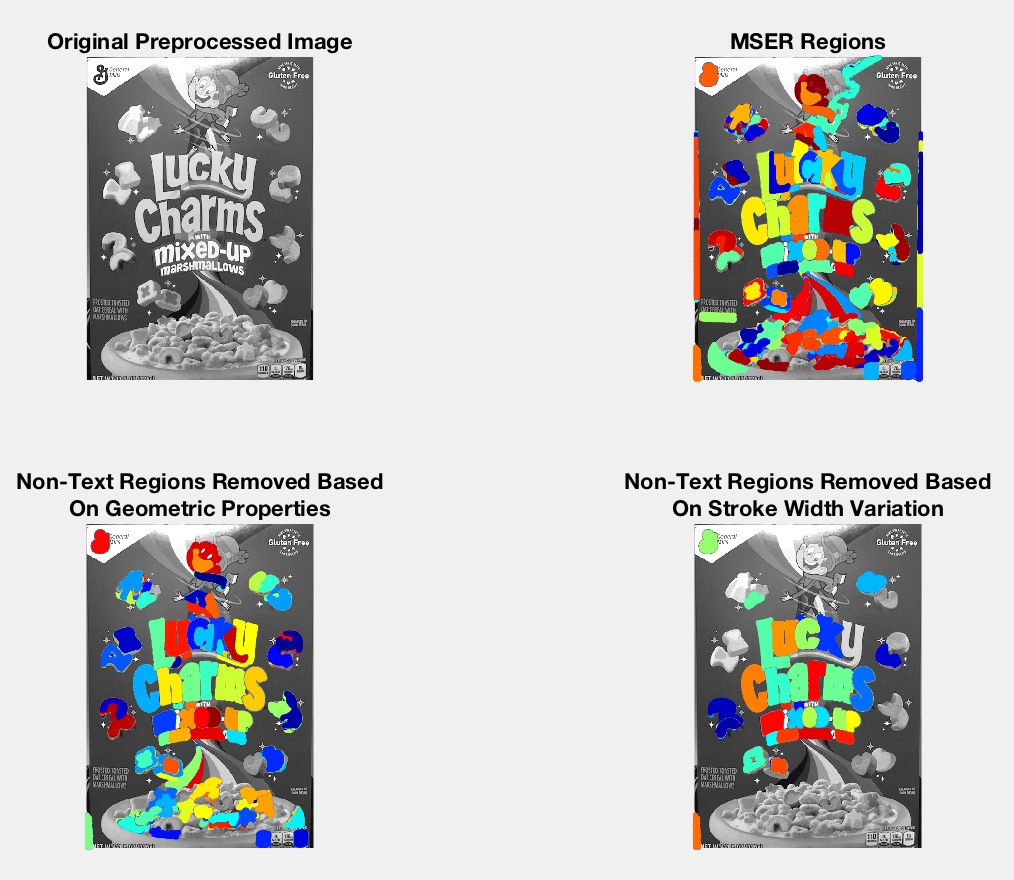}}
    \end{subfigure}
    \begin{subfigure}[]{\includegraphics[height=1.5in]{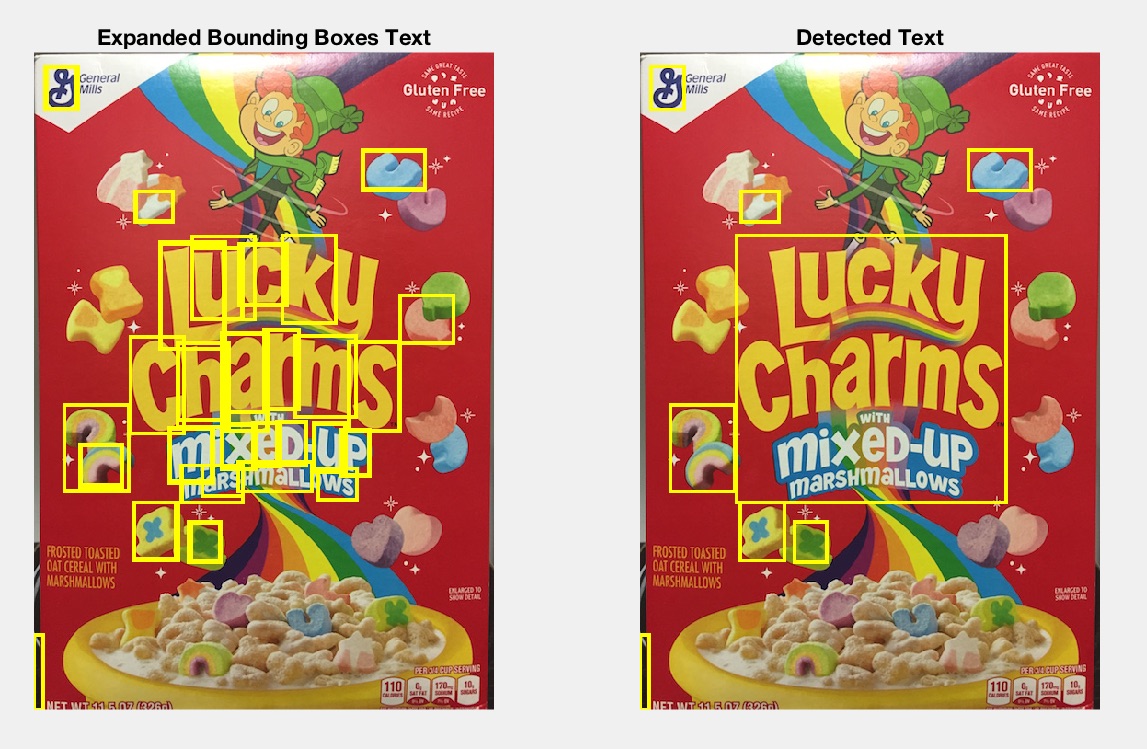}}
    \end{subfigure}
    \caption{Detected Features \& Text In Non-Natural Image}
\end{figure}

As expected though there are some discriminating geometric properties and other values that were more important for tuning to detect text in the non-natural images than in natural images all tuneable properties of this application were applicable in detecting text in the given test images.

\subsection{Issues With Detecting Text Using Custom, Highly Stylized Fonts}
In non-natural images there can be a lot variation among the subjects in an image that can affect text detection. Characteristics such as high contrast, line spacing and the spacing between the letter characters presented the most challenge for the approaches employed in the implemented text detection application. Despite these challenges the application with some tuning was able to successfully detect with reasonable proficient, primary and secondary text regions in the non-natural images.

The following geometric properties thresholds and related values that appear to have some sensitivity to high contrast images where text is highly stylized, custom fonts.  Note, these are not the only values that needed to be tuned, but these are the values that had to be tuned to process both non-natural images.

\subsubsection{Expansion Amount}
This value is used to tune amount by which the application expands neighboring bounding boxes. Neighboring bounding box expansion is an important step that is done in preparation of the final merge of individual character bounding boxes. Note, this final merge is were the bounding box is draw that suggests text candidates (words or lines of text). For, the non-natural images processed this value had to be lower which in terms increases the bounding box expansion. This was necessary in order to detect a wide and varied areas of text. Note, the size of the primary text regions in proportion to the rest of the subjects in the image could also contribute to this observation.

\subsubsection{Stroke Width Variation (Max) Threshold}
The value of the stroke width max threshold had to be increased by a factor of 2 in order to properly detect all individual text characters in the non-natural images vs the natural images.

\subsubsection{High Contrast Images}
One the pre-processing steps for most images is to increase the contrast of the image. For images that are by default high contrast images it was observed that increasing the image's contrast didn't contribute to better text detection performance, in fact in some instances the performance of text detection degraded. 

Note,the application implemented for this project uses a standard deviation based approach for performing image stretching (increasing image contrast).

\section{Future Work}
The next step is to automate the process of finding optimal threshold values for each image, for each region geometric property. By automating this process we will have a more robust and scalable means of conducting further analysis on different types of images and the properties that affected text detection the most. 

\section{Conclusion}
The approach of using Maximally Stable Extermal Regions (MSER) based feature detection, stroke width variation and geometric property thresholding even in a modest text detection application yields reasonably proficient results. Though highly stylized, custom fonts have some affect on the geometric properties thresholds used to filter non-text regions the affect was not as dramatic and did not have anticipated impact. This conclusion, reflects the power and robustness of all the techniques employed especially the use of MSER.

\ifCLASSOPTIONcaptionsoff
  \newpage
\fi

\bibliographystyle{IEEEtran}
\bibliography{mendeley}

\begin{thebibliography}{10}
\providecommand{\url}[1]{#1}
\csname url@samestyle\endcsname
\providecommand{\newblock}{\relax}
\providecommand{\bibinfo}[2]{#2}
\providecommand{\BIBentrySTDinterwordspacing}{\spaceskip=0pt\relax}
\providecommand{\BIBentryALTinterwordstretchfactor}{4}
\providecommand{\BIBentryALTinterwordspacing}{\spaceskip=\fontdimen2\font plus
\BIBentryALTinterwordstretchfactor\fontdimen3\font minus
  \fontdimen4\font\relax}
\providecommand{\BIBforeignlanguage}[2]{{%
\expandafter\ifx\csname l@#1\endcsname\relax
\typeout{** WARNING: IEEEtran.bst: No hyphenation pattern has been}%
\typeout{** loaded for the language `#1'. Using the pattern for}%
\typeout{** the default language instead.}%
\else
\language=\csname l@#1\endcsname
\fi
#2}}
\providecommand{\BIBdecl}{\relax}
\BIBdecl

\bibitem{Neumann2015a}
L.~Neumann and J.~Matas, ``{Efficient Scene text localization and recognition
  with local character refinement},'' \emph{Proceedings of the International
  Conference on Document Analysis and Recognition, ICDAR}, vol. 2015-Novem, pp.
  746--750, 2015.

\bibitem{Chen2011a}
H.~Chen, S.~S. Tsai, G.~Schroth, D.~M. Chen, R.~Grzeszczuk, and B.~Girod,
  ``{Robust text detection in natural images with edge-enhanced maximally
  stable extremal regions},'' \emph{Proceedings - International Conference on
  Image Processing, ICIP}, pp. 2609--2612, 2011.

\bibitem{Bergasa2012a}
L.~M. Bergasa and J.~J. Yebes, ``{Text Location in Complex Images},''
  \emph{International Conference on Pattern Recognition}, no. Icpr, pp.
  617--620, 2012.

\bibitem{Extremal2002}
M.~S. Extremal, J.~Matas, O.~Chum, M.~Urban, and T.~Pajdla, ``{Robust Wide
  Baseline Stereo from},'' \emph{In British Machine Vision Conference}, pp.
  384--393, 2002.

\bibitem{Republic}
C.~Republic, ``{FASText : Efficient Unconstrained Scene Text Detector}.''

\bibitem{Mikolajczyk2005}
K.~Mikolajczyk, T.~Tuytelaars, C.~Schmid, A.~Zisserman, J.~Matas,
  F.~Schaffalitzky, T.~Kadir, and L.~Van~Gool, ``{A comparison of affine region
  detectors},'' \emph{International Journal of Computer Vision}, vol.~65, no.
  1-2, pp. 43--72, 2005.

\bibitem{Republic2012a}
C.~Republic, ``{Real-Time Scene Text Localization and Recognition},'' pp.
  3538--3545, 2012.

\bibitem{Epshtein2010}
B.~Epshtein, E.~Ofek, and Y.~Wexler, ``{Detecting text in natural scenes with
  stroke width transform},'' \emph{Proceedings of the IEEE Computer Society
  Conference on Computer Vision and Pattern Recognition}, no.~d, pp.
  2963--2970, 2010.

\bibitem{coco2014}
\BIBentryALTinterwordspacing
``{COCO Common Objects}.'' [Online]. Available: \url{http://mscoco.org/home/}
\BIBentrySTDinterwordspacing

\bibitem{tessORC}
\BIBentryALTinterwordspacing
``{Tesseract Open Source OCR Engine}.'' [Online]. Available:
  \url{https://github.com/tesseract-ocr/tesseract}
\BIBentrySTDinterwordspacing

\end{thebibliography}

\end{document}